\documentclass[runningheads]{llncs}
\usepackage{graphicx}
\usepackage{comment}
\usepackage{amsmath,amssymb} % define this before the line numbering.
\usepackage{color}
\usepackage{paralist}
\usepackage{subfigure}
\usepackage{stackengine}
\usepackage{adjustbox}
\usepackage{caption} 
\usepackage{multirow}
\usepackage{diagbox}
\usepackage[colorlinks,linkcolor=blue]{hyperref}
\usepackage{bbm}
\usepackage{booktabs}
\usepackage{bm}
\usepackage{bbding}

\usepackage[colorlinks,linkcolor=blue]{hyperref}
\begin{document}
\title{CLIP-DR: Textual Knowledge-Guided Diabetic Retinopathy Grading with Ranking-aware Prompting}
\titlerunning{CLIP-DR for Diabetic Retinopathy Grading}
\author{
Qinkai Yu \inst{1}\and
Jianyang Xie\inst{2}\and
Anh Nguyen\inst{3}\and
He Zhao\inst{2}\and
Jiong Zhang\inst{4}\and
Huazhu Fu\inst{5}\and
Yitian Zhao\inst{4} and
Yalin Zheng\inst{2}\and
Yanda Meng \inst{1(}\Envelope\inst{)}}

\authorrunning{Q.Yu et al.}
\institute{
\textsuperscript{$1$} Computer Science Department, University of Exeter, Exeter, UK\\
\textsuperscript{$2$} Eye and Vision Sciences Department, University of Liverpool, Liverpool, UK\\
\textsuperscript{$3$} Computer Science Department, University of Liverpool, Liverpool, UK\\
\textsuperscript{$4$} Ningbo Institute of Materials Technology and Engineering, Chinese Academy of Sciences, Ningbo, China \\
\textsuperscript{$5$} Institute of High Performance Computing, Agency for Science, Technology and Research (A*STAR), Singapore. \\
\email{Y.M.Meng@exeter.ac.uk} \\
}
%index{Qinkai Yu}
%index{Jianyang Xie}
%index{Anh Nguyen}
%index{Jiong Zhang}
%index{Huazhu Fu}
%index{Yitian Zhao}
%index{Yalin Zheng}
%index{Yanda Meng}
% If the paper title is too long for the running head, you can set
% an abbreviated paper title here
%
%s
%\authorrunning{}
% First names are abbreviated in the running head.
% If there are more than two authors, 'et al.' is used.
\maketitle
\begin{abstract}
Diabetic retinopathy (DR) is a complication of diabetes and usually takes decades to reach sight-threatening levels. Accurate and robust detection of DR severity is critical for the timely management and treatment of diabetes. However, most current DR grading methods suffer from insufficient robustness to data variability (\textit{e.g.} colour fundus images), posing a significant difficulty for accurate and robust grading. In this work, we propose a novel DR grading framework CLIP-DR based on three observations: 1) Recent pre-trained visual language models, such as CLIP, showcase a notable capacity for generalisation across various downstream tasks, serving as effective baseline models. 2) The grading of image-text pairs for DR often adheres to a discernible natural sequence, yet most existing DR grading methods have primarily overlooked this aspect. 3) A long-tailed distribution among DR severity levels complicates the grading process. This work proposes a novel ranking-aware prompting strategy to help the CLIP model exploit the ordinal information. Specifically, we sequentially design learnable prompts between neighbouring text-image pairs in two different ranking directions. Additionally, we introduce a Similarity Matrix Smooth module into the structure of CLIP to balance the class distribution. Finally, we perform extensive comparisons with several state-of-the-art methods on the GDRBench benchmark, demonstrating our CLIP-DR's robustness and superior performance. The implementation code is available \footnote{\url{https://github.com/Qinkaiyu/CLIP-DR}}.
\end{abstract}
\section{Introduction}
\begin{figure*}[tb]
\centering
\includegraphics[width=9cm]{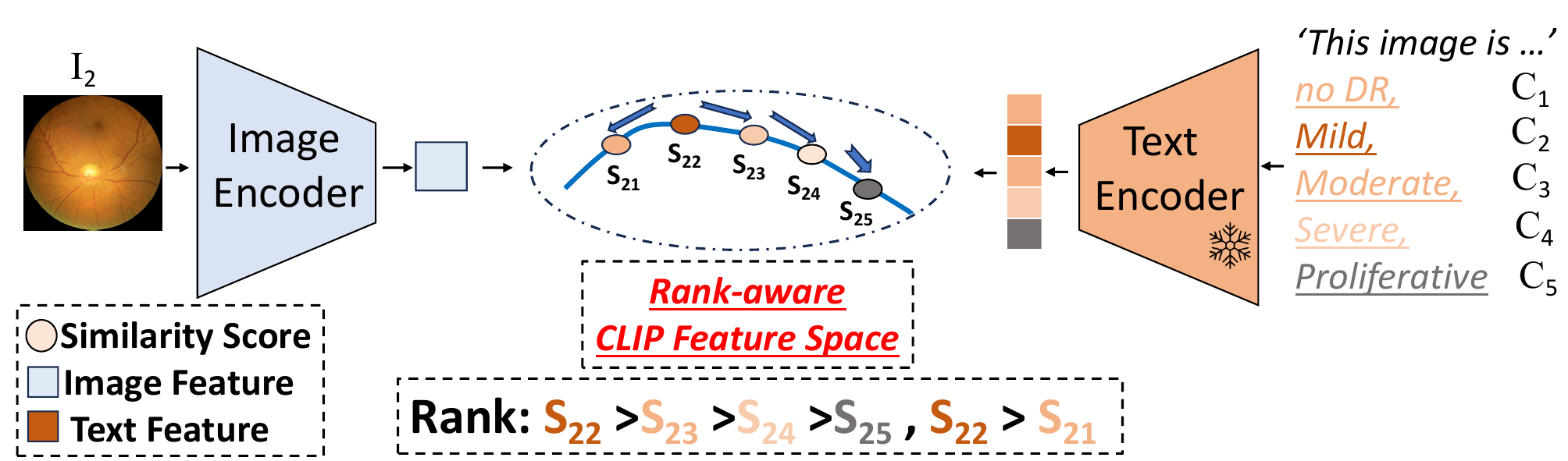}
\caption{\textbf{An example of learnable rank-aware prompting with image class `Mild' for a given Image $I_{2}$.} $[C_{1},...,C_{5} ]$ represent 5 different DR grading classes. The similarity score is obtained by the inner product of the image feature and text feature. Designing learnable rank-aware prompts that satisfy the following two inequalities enables the model to learn natural order information.}
\label{figure1}
\end{figure*}

Diabetic retinopathy (DR) is a main complication of diabetes, typically progressing over several years before reaching levels that threaten vision. The disease process can present a wide range of severity grades and change over time. These grades of severity are often categorised into different classes (\textit{e.g.}, normal, mild, moderate, severe \cite{kempen2004prevalence}), and the variations among these classes are often difficult to discern. Stylistic appearance variability in the different data sources of colour fundus images further complicates this challenge \cite{li2020siamese}. 
%Although deep learning is capable of tackling a variety of classification tasks\cite{lecun1998gradient,he2016deep,radford2021learning}, classification of DR grade is still challenging\cite{porwal2020idrid}.
A growing number of deep learning studies researched the DR grading tasks with colour fundus images \cite{atwany2022drgen,zhang2017mixup,zhou2021domain,liu2020green,he2020cabnet,zhou2020deep}. However, most of these previous studies attribute the poor performance of classification models in DR grading to the diversity of diagnostic patterns, data imbalance, and large differences in individual appearance styles. Recently, the Contrastive Language-Image Pre-Training (CLIP) model \cite{radford2021learning} has gained significant attention for its strong generalisation capability in various downstream vision tasks, particularly achieving high classification accuracy for unbalanced or stylistically diverse data. Many paradigms of CLIP in the medical domain have also shown great potential. For instance, CLIP performance was enhanced by prompt strategies for combining medical knowledge \cite{wu2023medklip}. In this work, we propose a novel framework (an example is shown in Fig. \ref{figure1}), CLIP-DR, that takes advantage of CLIP's robust and sufficient feature learning ability. %An example is shown in Fig. \ref{figure1}.

Most DR grading studies \cite{yang2021adversarial,rame2022fishr,yang2022multi,che2023towards,kempen2004prevalence} assumed that DR labels were independent.
%, and classified the labels into five different groups~\cite{kempen2004prevalence}: no DR, mild, moderate, severe and proliferation. 
The different groups' ground truth was converted to a one-hot format during training. However, the fact that the grade of DR follows an underlying natural order was ignored. The commonly adopted loss function of Cross-Entropy by previous methods \cite{wang2022image} resulted in the same penalty regardless of the category to which the sample was misclassified. Typically, for the tasks with a natural order such as age estimation \cite{niu2016ordinal}, the regular approach is to treat classification as a metric regression problem \cite{fu2008human}, minimising the absolute/squared error loss (\textit{i.e.}, MAE/MSE). However, by treating discrete data as continuous values, the regression models are prone to ambiguity in the boundaries between different classes, making it difficult to distinguish between neighbouring classes \cite{wang2023ord2seq}. Therefore, it is not appropriate to approach DR grading purely from a classification or regression point of view.

\begin{figure*}[tb]
    \centering
{\includegraphics[width=11cm]{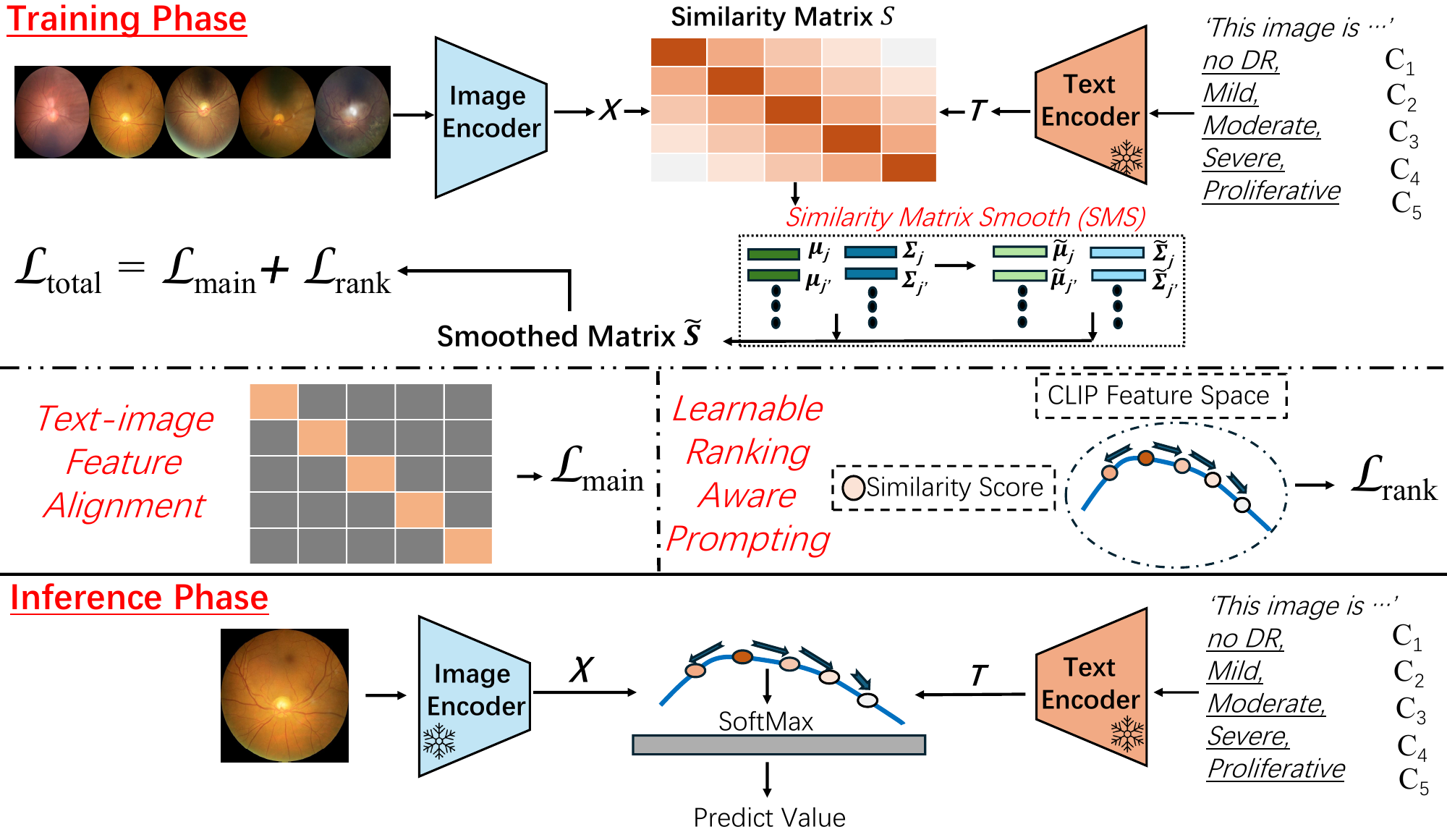}}
    \caption{Overview of the proposed CLIP-DR framework for training and inference. 
    %It contains an image encoder, text encoder and similarity matrix smoothing (SMS) modules. 
    Images are processed through an image encoder to extract image features $X$. The corresponding text labels are fed into the text encoder, generating text embeddings for labels $T$. The similarity matrix $\mathcal{S}$ is obtained through the inner product. Finally, the SMS module converts $\mathcal{S}$ into calibration features $\tilde{\mathcal{S}}$ with the same dimensions. 
    The learnable rank-aware prompt strategy is implemented explicitly by $\mathcal{L}_{rank}$, which uses ranking information independently in the left and right directions, and $\mathcal{L}_{main}$ follows the practice of CLIP \cite{radford2021learning}.
    }
    \label{fig2}
    \vspace{-5pt}
\end{figure*}

The challenges above lead to the question: \textbf{Can we think out of the box and rationalise the fact that DR conforms to the natural order to improve grading performance?}  We propose a new CLIP-based framework (CLIP-DR, pipeline shown in Fig. \ref{fig2}) treating DR grading as an image-text matching problem. Additionally, we propose a ranking-aware prompting strategy to fine-tune the CLIP image encoder. This enables CLIP-DR to learn the associations of natural ordering information under a classification task. Specifically, we minimise the Kullback-Leibler (KL) divergence as the primary loss function. The ranking loss function is proposed as ranking-aware prompting, empowering the image encoder to grasp natural ordering information. Its purpose is to ensure that the ranking aligns consistently with the inherent order of DR. We further improve the grading performance by introducing a Similarity Matrix Smooth (SMS) module into the framework to minimise the impacts due to data imbalance/long-tailed distribution. We compare the performance of CLIP-DR with other state-of-the-art methods on benchmark GDRBench \cite{che2023towards}. Experimental results demonstrate CLIP-DR's effectiveness and superior performance.
\section{Methods}

%In this section, we first introduce our CLIP-DR framework. Next, we explain in detail that we propose an effective loss function to help the CLIP model learn the ranking information among the data. In addition, we introduce the FDS module to minimize the effect of data imbalance. The details of our framework are shown in Fig\ref{fig2}.\\
\subsection{Problem Formulation}
Given an input colour fundus image $I_{i}$, and its corresponding class $C_{i}$, a pair of them $\{I_{i}, C_{i}\}$ is an element of DR grading dataset $\mathcal{D}$. Suppose there is $K$ number of different classes; the set $\mathcal{T} = \{t_{j}\}^{K}_{j=1}$ can be defined as a set of text embeddings of classes $C$. We also define $x_{i}$ as the image embeddings of $I_{i}$ from the image encoder and set $\mathcal{X} = \{x_{i}\}^{M}_{i=1}$. Where $M$ is the number of images. With CLIP, text and images can be mapped to the same dimension in our framework, such as $x_{i}, t_{j} \in \mathbb{R}^{1\times1024}$. The similarity matrix $\mathcal{S}$ between set $\mathcal{X}$ and $\mathcal{T}$ is calculated via inner product, where $\mathcal{S} = [s_{i,j}]_{M\times K}\in \mathbb{R}^{M\times K}$, and $s_{i,j} = x_{i} \cdot t_{j}^{T}$.
% \begin{equation}
%     s_{i,j} = x_{i} \cdot t_{j}^{T}
% \end{equation}
Then for a given sample $\{I_{i},C_{i}\}$, our aim is to learn a mapping $f_{\theta}: I_{i} \rightarrow C_{i}$, where $f$ is a deep neural network with parameters $\theta$.\\

%Our framework is shown in Fig \ref{fig2}. Images are processed through an image encoder to extract image features $\mathcal{X}$. The corresponding text labels are fed into the text encoder, generating text embeddings for labels $\mathcal{T}$. The similarity matrix $\mathcal{S}$ is obtained through the inner product. Finally, the SMS module converts $\mathcal{S}$ into calibration features $\tilde{\mathcal{S}}$ with the same dimensions. 
% The learnable rank-aware prompt strategy is implemented explicitly by the loss function $\mathcal{L}_{rank}$, which uses ranking information independently in the left and right directions, and $\mathcal{L}_{main}$ following the practice of CLIP \cite{radford2021learning}.

\subsection{SMS for Imbalanced Regression}
This section presents the Similarity Matrix Smooth (\textit{SMS}) module. We integrate SMS into CLIP-DR by inserting a feature calibration layer after the similarity matrix. 
In essence, SMS is a mapping $ g: \mathbb{R}^{M\times K} \rightarrow \mathbb{R}^{M\times K}: \mathcal{S} \rightarrow \tilde{\mathcal{S}}$ designed to smooth the original similarity matrix. This smoothing process is inspired by \cite{yang2021delving} and aims to mitigate the impact of data imbalance, with the resulting smoothed statistic reflecting the ordinal relationship among neighbouring targets. Our SMS addresses data imbalance by smoothing similarity vectors, while \cite{yang2021delving} targets entropy.
We define $s_{.,j} \in \mathbb{R}^{1\times K}$ as the row vector that the true text embedding is $t_{j}$. 
By estimating the statistics of $s_{.,j}$, we can easily obtain the mean and variance. Then we define $\mu_{j}$ as the mean of all row vectors $s_{.,j}$ and $\Sigma_{j}$ as the variance. After employing a symmetric kernel
$ \mathcal{K}$ to smooth the distribution of $\mu_{j}$ and $\Sigma_{j}$, they can be defined as:

\begin{equation}
    \tilde{\mu}_{j} = \sum_{j'\neq j}^{K} \mathcal{K}(Y_{\cdot,j},Y_{\cdot,j'})\mu_{j'}, \quad \tilde{\Sigma}_{j} = \sum_{j'\neq j}^{K} \mathcal{K}(Y_{\cdot,j},Y_{\cdot,j'})\Sigma_{j'}
\end{equation}
Then we calibrate the similarity vector $s_{.,j}$ into $\tilde{s}_{.,j}$, such as:
\begin{equation}
    \tilde{s}_{.,j} = \tilde{\Sigma}_{j}^{\frac{1}{2}} \tilde{\mu}_{j}^{\frac{1}{2}}(s_{.,j}-\mu_{j})+\tilde{\mu}_{j}
\end{equation}
$\tilde{\mu}_{j}$ and $\tilde{\Sigma}_{j}$ updated across different epochs but fixed within each training epoch. Finally,  stacking $ \tilde{s}_{.,j}$ by rows yields $\tilde{\mathcal{S}}$. Such a smoothed matrix $\tilde{\mathcal{S}}$ can calibrate potentially biased estimates of similarity matrix distributions, especially for classes with few samples, thus mitigating the impact of data imbalance for prompt learning.
\subsection{Prompting Rank-aware Gradient}
The core idea of rank-aware prompting is to exploit the natural order of DR colour fundus images, thus improving the grading accuracy. The motivation comes from a reasonable explanation: misdiagnosis of the highest-ranked patient as having no disease is more severe than misdiagnosis of the highest-ranked patient as having an intermediate disease. The details of rank-aware gradient prompting are elaborated below. 

\noindent  \textbf{Main Loss.} Our main loss consists of image-to-text loss and text-to-image loss. Recall that if $t_{j}$ is the text embeddings corresponding to $I_{i}$, then $s_{i,j}$ is the similarity score of this correct matching. $\mathcal{S} \in \mathbb{R}^{M\times K}$ is a text-to-image pair matrix, while $\mathcal{S}^T \in \mathbb{R}^{K\times M}$ is a image-to-text pair matrix. The SMS module converts $\mathcal{S}$ to $\tilde{\mathcal{S}}$. Thus the corresponding label can be expressed as $Y$ for text-to-image pair in Eq. \ref{indicator1} and $Y^T$ for text-to-image pair:
\begin{equation}
\label{indicator1}
    Y = \mathbb{I}(\tilde{\mathcal{S}})=\left\{
\begin{aligned}
&1,\: \text{correct matching}\\
&0, \: \text{otherwise}\\
\end{aligned}
\right.
\end{equation}

%Since $K\leq M$ and $\tilde{\mathcal{S}}$ is not a square matrix, the image-to-text label ($Y^{T}$) may have zero or multiple hits, It is inappropriate to consider similarity score learning as a 1-in-N classification problem with cross-entropy loss along the lines of original CLIP \cite{wang2021actionclip}. 
Given that $K\leq M$ and $\tilde{\mathcal{S}}$ is not square, the image-to-text label ($Y^{T}$) could potentially yield zero or multiple hits. Consequently, treating similarity score learning as a 1-in-N classification problem with cross-entropy loss, akin to the original CLIP framework \cite{wang2021actionclip}, would be unsuitable. Thus, inspired by previous methods \cite{wang2021actionclip,li2022ordinalclip}, we adopt the \textit{KL} divergence loss. Specifically, we
%In this case, it is common to define the Kullback-Leibler (KL) divergence as loss function\cite{}. We 
construct the new label matrix $Y'$ using the non-zero columns of the normalised transpose of $Y^{T}$. $\tilde{\mathcal{S}}^T$ is also applied via the softmax layer to obtain the normalised matrix $\tilde{\mathcal{S'}}^T$. Therefore, image-to-text loss and text-to-image loss can be expressed as:
\begin{equation}
\text{image-to-text: } \frac{1}{K} \sum_{i=1}^{K}KL(Y'_{i,\cdot}\|  \tilde{\mathcal{S'}}_{i, \cdot}^T), \quad \text{text-to-image: }\frac{1}{M} \sum_{i=1}^{M}KL(Y_{i,\cdot}\|  \tilde{\mathcal{S'}}_{i, \cdot})
\end{equation}
Note that the normalised $Y$ is equal to itself since there is only one hit in each row and $\tilde{\mathcal{S'}}$ is the normalised matrix by $\tilde{\mathcal{S}}$.
At the end, the main loss is defined in Eq. \ref{main loss} as:
\begin{equation}
\label{main loss}
   \mathcal{L}_{main} =  \frac{1}{2}[\frac{1}{K} \sum_{i=1}^{K}KL(Y'_{i,\cdot}\|  \tilde{S'}_{i, \cdot}^T)+  \frac{1}{M} \sum_{i=1}^{M}KL(Y_{i,\cdot}\|  \tilde{S'}_{i, \cdot})]
\end{equation}

\noindent  \textbf{Rank Loss.} For a given image $I_{i}$, $t_{j}$ is its corresponding text embeddings and $s_{i,j}$ is the similarity score of this correct matching. The DR coloured fundus of the overall dataset has a natural grade of information, although very few patients might experience, for example, a direct transition from mild or moderate DR to proliferative DR, without experiencing severe DR. We aim to prompt the rank information of DR grading by ensuring that:

\begin{equation}
\label{eq7}
  \textcolor{black}{
\left\{ \begin{array}{ll}
    \tilde{s}_{i,j}>\tilde{s}_{i,j+1}>...>\tilde{s}_{i,K} \\
    \tilde{s}_{i,j}>\tilde{s}_{i,j-1}>...>\tilde{s}_{i,1}
\end{array} \right.}
\end{equation}
To this end, we propose a novel rank-loss that performs binary classification for each neighbouring class in two directions (left and rightward). For example, for a neighbouring class pair $(\tilde{s}_{i,j'},\tilde{s}_{i,j'+1})$, we design the loss function by minimising the gap between this pair and the label (1, 0). Thus, the rank loss function ($\mathcal{L}_{\text{rank}}$) is defined in Eq. \ref{eq:rank_loss} as:
\begin{equation}
\label{eq:rank_loss}
    \mathcal{L}_{\text{rank}} = -\frac{1}{M}\sum_{i=1}^{M}( \mathcal{L}_{\text{rightward}}^{i} +  \mathcal{L}_{\text{leftward}}^{i})
\end{equation}
where $\mathcal{L}_{\text{rightward}}^{i}$ and $\mathcal{L}_{\text{leftward}}^{i}$ are the learnable prompt approach designed to fit the goals in Eq. \ref{eq7}. They are defined as:
\begin{equation}
    \mathcal{L}_{\text{rightward}}^{i} =  -{\sum_{j'= j}^{k-1}\log \frac{\exp(\tilde{s}_{i,j'}/ \tau)}{\exp(\tilde{s}_{i,j'}/ \tau)+\exp(\tilde{s}_{i,j'+1}/ \tau)}},
\end{equation}
\begin{equation}
    \mathcal{L}_{\text{leftward}}^{i} =  -\sum_{j'=2}^{j}\log \frac{\exp(\tilde{s}_{i,j'}/ \tau)}{\exp(\tilde{s}_{i,j'}/ \tau)+\exp(\tilde{s}_{i,j'-1}/ \tau)},
\end{equation}
where we set label of $\tilde{s}_{i,j'}$, $\tilde{s}_{i,j'+1}$ and $\tilde{s}_{i,j'-1}$ is 1, 0, 0, respectively. The $\tau$ was set to 1 during the experiment. It is worth noting that although the proposed $\mathcal{L}_{\text{rank}}$ is similar to the kappa loss \cite{de2018weighted} (the kappa loss is usually used for classification of ordered labels), unlike fixed weights in kappa loss which are linear or quadratic, the weights in $\mathcal{L}_{\text{rank}}$ are in a self-adaptive non-linear fashion, making the $\mathcal{L}_{\text{rank}}$ applicable to the learnable prompting text-image alignment in clip feature space.
In the end, the total loss is defined as $\mathcal{L}_{\text{total}} = \mathcal{L}_{\text{main}}+ \lambda \mathcal{L}_{\text{rank}}$,
% \begin{equation}
%     \mathcal{L}_{total} = \mathcal{L}_{main}+ \lambda \mathcal{L}_{rank}
% \end{equation}
where $\lambda$ is a hyper-parameter weighting the loss term $\mathcal{L}_{\text{rank}}$, which is set as 1 by default.
\section{Datasets and Implementation Details}
We follow the same experiment setting of GDRBench \cite{che2023towards} involving two generalization ability evaluation settings and eight popular public datasets.\\
\textbf{Evaluation Settings}. The initial evaluation follows the classic leave-one-domain-out protocol (DG test), where one domain is withheld for evaluation while models are trained on the remaining domains. The DG test encompasses six datasets: DeepDR \cite{zeng2019deepdr}, Messidor \cite{decenciere2014feedback}, IDRID \cite{porwal2020idrid}, APTOS \cite{karthick2019aptos}, FGADR \cite{zhou2020benchmark}, and RLDR \cite{wei2021learn}. Another evaluation scenario is the extreme single-domain generalization setting (ESDG test), which adopts a train-on-single-domain protocol using the aforementioned datasets.
%, evaluates on two extra large-scale datasets, DDR \cite{li2019diagnostic} and EyePACS \cite{EYEPACS2023}. 
We adopted two essential metrics to evaluate the performance: the area under the ROC curve (AUC) and macro F1-score (F1).  We used \textbf{bold} and \underline{underline} to indicate the first and the second-highest scores in each sub-dataset test performance. \\
\noindent \textbf{Implementation Details}. All experiments are conducted on an Nvidia 4090 GPU. CLIP \cite{radford2021learning}, OrdinalCLIP \cite{li2022ordinalclip}, and our CLIP-DR used pre-trained ResNet50 \cite{he2016deep} as the image encoder backbone and text encoder is a pre-trained Transformer. The initial prompt was \textit{``This image is \{label\}"}.  The number of training epochs is set to 100.
\section{Results}
\begin{figure*}[tb]
\centering
{\includegraphics[width=1\linewidth]{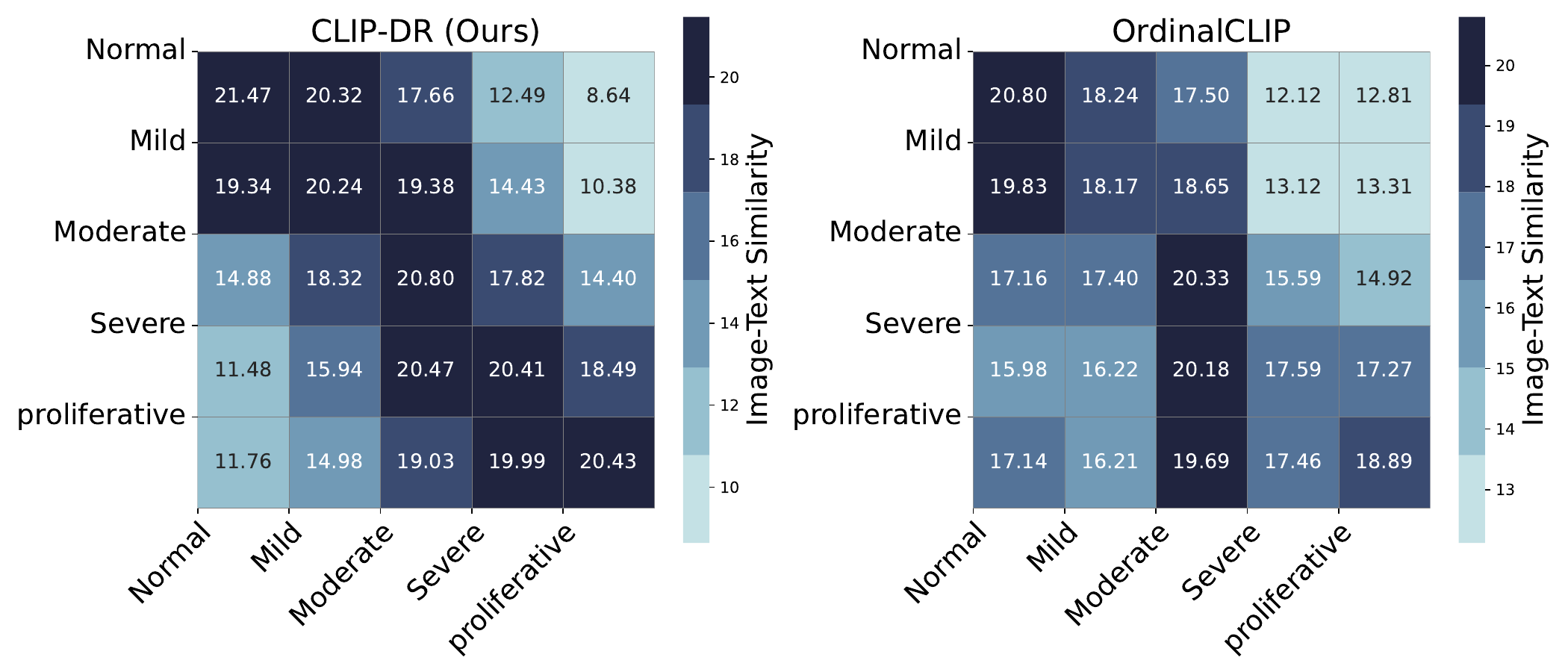}}
\caption{The image-text similarity matrix obtained by the inner product of image feature and text feature: $X\cdot T^{t}$. This matrix is an intuitive representation of the rank between different image-text pairs. The X-axis represents five different text labels, and the Y-axis represents real images of different classes. 
%Notably, CLIP-DR only fuzzily classifies a few neighbouring classes, except for the 'Severe' label image, which is easily predicted to be 'Moderate'.
We average the results of the six sub-datasets and present them in this figure for our CLIP-DR and OrdinalCLIP \cite{li2022ordinalclip}. CLIP-DR can learn rank-aware text-image features.}
\label{hot}
\end{figure*}

% \begin{figure*}
% \centering
% \includegraphics[width=9cm]{Figures/hot.pdf}
% \caption{The image-text similarity matrix obtained by the inner product of image feature and text feature: $X\cdot T^{t}$. This matrix is an intuitive representation of the rank between different image-text pairs. The X-axis represents 5 different text labels, and the Y-axis represents real images of different classes. 
% %Notably, CLIP-DR only fuzzily classifies a few neighbouring classes, except for the 'Severe' label image, which is easily predicted to be 'Moderate'.
% We average the results of the six sub-datasets and present them in this figure.}
% \label{hot}
% \end{figure*}

\begin{table*}[tb]
\caption{Comparison with state-of-the-art approaches under the DG test.}
\renewcommand\arraystretch{1.5}
\centering 
\scalebox{0.7}{
\begin{tabular}{l|cc|cc|cc|cc|cc|cc|cc}
\hline
Target & \multicolumn{2}{|c|}{APTOS} & \multicolumn{2}{|c|}{DeepDR} & \multicolumn{2}{|c|}{FGADR} &\multicolumn{2}{|c|}{IDRID}&\multicolumn{2}{|c|}{Messidor}&\multicolumn{2}{|c|}{RLDR}&\multicolumn{2}{|c}{Average}\\
\hline

Metrics&F1& AUC  & F1  & AUC  & F1 & AUC &F1&AUC&F1&AUC&F1&AUC&F1&AUC \\
\hline
\hline
DRGen \cite{atwany2022drgen} & 40.2&\underline{79.9}   & 34.1& 83.0   &24.7& 69.4 &37.4&\textbf{84.7}&40.5&79.0&37.0&\underline{79.5}&37.3&79.3  \\
Mixup \cite{zhang2017mixup} & 43.2& 75.3 & 25.2  & 75.3   &32.3& 66.7 &27.6&78.8&32.6&76.7&37.7&76.9&33.1&75.0 \\
MixStyle \cite{zhou2021domain} & 39.9& 79.0   & 27.9 & 76.9  &22.7& 71.2 &\underline{39.2}&83.0&36.5&75.2&31.4&75.5&32.9&76.8 \\
GREEN \cite{liu2020green}  & 38.9 & 75.1  & 24.9 & 76.4 &31.5& 69.5 &32.2&79.9&36.8&75.8&34.4&74.8&33.1&75.3 \\
CABNet \cite{he2020cabnet} & 39.4& 75.8  & 31.8 & 75.2  &34.8 & 73.2 &37.3&79.2&34.1&74.2&35.6&75.8&35.5&75.6 \\
DDAIG \cite{zhou2020deep} & 41.0& 78.0  & 32.2 & 75.6  &33.8& 73.6 &27.0&82.1&35.3&76.6&27.7&75.6&32.8&76.9 \\
ATS \cite{yang2021adversarial}&38.3&77.1&31.6&79.4&33.4&74.7&34.9&83.0&35.8&77.2&34.9&76.5&34.8&78.0\\
Fishr \cite{rame2022fishr}&43.4&79.2&34.4&81.1&34.4&73.3&27.6&82.7&41.1&76.4&34.7&77.4&35.9&78.4\\
MDLT \cite{yang2022multi}&41.5& 77.3&36.2&80.0&29.0&74.1&35.4&81.5&36.9&75.4&35.0&75.7&35.7&77.3\\
GDRNet \cite{che2023towards} &\underline{46.0}& 79.9 &\underline{45.3}&84.7&\underline{39.4}&\textbf{80.8}&35.9&84.0&\textbf{50.9}&\textbf{83.2}&\textbf{43.5}&\textbf{82.9}&\underline{43.5}&\underline{82.6}\\
\hline
CLIP \cite{radford2021learning} &44.3&76.1 &42.0&83.6&41.1&78.6&34.8&83.0&39.6&76.7&38.8&75.9&40.1 & 78.9\\
OrdinalCLIP \cite{li2022ordinalclip} &45.7&77.6 &43.3&\underline{85.1}&37.9&79.3&36.2&80.4&41.8&77.7&39.5&76.6&40.7&79.4\\
CLIP-DR(Ours)&\textbf{46.3}&\textbf{83.3}&\textbf{45.8}&\textbf{89.9}&\textbf{48.0}&\underline{80.7}&\textbf{41.9}&\underline{84.5}&\underline{47.3}&\underline{79.0}&\underline{41.0}&78.9&\textbf{45.5}&\textbf{82.7}\\
\bottomrule[1pt]
\end{tabular}
\label{tab:main-results}
}
\end{table*}

As shown in Table \ref{tab:main-results}, We compare the proposed method with the following state-of-the-art methods: DRGen \cite{atwany2022drgen}, Mixup \cite{zhang2017mixup}, MixStyle \cite{zhou2021domain}, GREEN \cite{liu2020green}, CABNet \cite{he2020cabnet}, DDAIG \cite{zhou2020deep}, ATS \cite{yang2021adversarial}, Fishr \cite{rame2022fishr}, MDLT \cite{yang2022multi}, GDRNet \cite{che2023towards}, CLIP \cite{radford2021learning} and OrdinalCLIP \cite{radford2021learning}. 
%It is notable that RETFound is a recently introduced large model for retinal images that has achieved excellent results in DR grading. To the best of our knowledge, its training data contains 160w+ images, which include well-known public datasets such as IDRiD and Messidor. Therefore it is unfair to do DG test and ESDG test using RETFound. 
Table \ref{tab:main-results} shows the results of the DG test setting, where the target row indicates the test set. 
%Our proposed CLIP-DR outperforms the other methods, and since 
Our CLIP-DR achieved the \textbf{best} performance of the F1 score on four sub-tests datasets and the second-best F1 score performance on the other two sub-tests datasets. On average, our CLIP-DR achieved the \textbf{best} performance of \textit{F1} and \textit{AUC} across all six datasets. Our model outperformed GDRNet\cite{che2023towards} (SOTA) with a statistically significant p-value of 0.02, where ours achieved the best F1 or AUC.
Note that Messidor \cite{decenciere2014feedback} and RLDR \cite{wei2021learn} contain excessive differences in colour styles from the other datasets \cite{che2023towards}.
GDRNet \cite{che2023towards} specially designed strong data enhancement techniques for finite-domain transformations to address such a challenge and thus achieved good performance. 
Notably, OrdinalCLIP \cite{li2022ordinalclip} is closely related work that exploits the ordinal information when tackling regression tasks. OrdinalCLIP derives ordinal ranking embeddings $R \in \mathbb{R}^{K\times 1024}$ from a set of basic ranking embeddings $R'\in \mathbb{R}^{K'\times 1024}$ by interpolation, and the base ranking embedding length $K'$ needs to be much smaller than the ordinal ranking embeddings length $K$ (\textit{e.g.,} $K' \ll K$). However, there are only five classes in DR grading, so finding such a base ranking embedding length is hard. Both our approach and OrdinalCLIP \cite{li2022ordinalclip} aim to learn ranked feature spaces in CLIP \cite{radford2021learning}. While OrdinalCLIP \cite{li2022ordinalclip} uses linear interpolation, we make DR fundus order learnable via text-image pairs.
Compared to CLIP \cite{radford2021learning} and OrdinalCLIP \cite{li2022ordinalclip}, our CLIP-DR significantly improves all sub-test datasets' grading performance. 
A heat map of the similarity matrix of OrdinalCLIP \cite{li2022ordinalclip} and our CLIP-DR averaged over the 6 subtests is shown in Fig. \ref{hot}. It shows that 
our CLIP-DR satisfies the objectives in Eq. \ref{eq7}, where a rank-aware text-image feature space is learned. While OrdinalCLIP \cite{li2022ordinalclip} cannot reflect such rank information. To be more explicit about the effectiveness of our approach compared to OrdinalCLIP, we give the class activation map for CLIP-DR and OrdinalCLIP in the appendix.
% This demonstrates the robustness of our proposed rank-aware prompting. 
We also provide the AUC performance for each class under the DG test setting in the appendix for comprehensive experimental results. 
When lacking training examples, the previously widely used ``pretraining-finetuning" paradigm would fail to fine-tune the entire CLIP backbone \cite{gao2024clip}. 
% However, fine-tuning a pre-trained visual-language model (\textit{e.g.} CLIP) needs a relatively large number of data, as proven before \cite{gao2024clip}. 
% Thus, our CLIP-DR falls short in this particular ESDG test setting. However, to conduct comprehensive experiments, we evaluate CLIP-DR in the ESDG setting and the results can be found in the supplementary material.
ESDG experiment setting with little training data cannot exploit the benefit of pre-trained CLIP. The intuition of this work is to boost the CLIP performance through prompt learning by exploiting the ranking information of DR images with relatively sufficient data. However, we still experimented with ESDG test setting for comprehensive experimental results; the comparison results can be found in the appendix.
%The average performance of CLIP-DR on the DG test reaches SOTA. 

\noindent \textbf{Ablation study of proposed components.}
\begin{table*}[tb]
\caption{Ablation studies under the DG test.}
\renewcommand\arraystretch{1.5}
\centering
\scalebox{0.8}{%
\begin{tabular}{l|cc|cc|cc|cc|cc|cc|cc}
\hline
\multirow{1}{*}{Target} & \multicolumn{2}{|c|}{APTOS} & \multicolumn{2}{|c|}{DeepDR} & \multicolumn{2}{|c|}{FGADR} &\multicolumn{2}{|c|}{IDRID}&\multicolumn{2}{|c|}{Messidor}&\multicolumn{2}{|c|}{RLDR}&\multicolumn{2}{|c}{Average}\\
%\cmidrule(lr){2-6}
%\cmidrule(lr){7-7}
%\cmidrule(lr){8-9}
\hline

Metrics&F1& AUC  & F1  & AUC & F1 & AUC &F1&AUC&F1&AUC&F1&AUC&F1&AUC \\
\hline
\hline
\textit{w/o} $\bm{\mathcal{L}_{\text{main}}}$&44.0&82.2&40.5&86.5&33.4&80.5&37.4&83.8&45.7&78.7&38.1&78.3&39.85&81.6\\
\textit{w/o} $\bm{\mathcal{L}_{\text{rank}}}$&45.7& 75.7&43.0&84.3&37.4&77.8&36.9&79.5&42.0&76.9&39.4&76.2&40.88&78.4\\

\textit{w/o} \textit{SMS}&45.5&  80.9 &44.9&88.2&45.6&79.9&40.0&82.9&46.4&78.1&40.3&77.0&43.78&81.1\\
\hline
CLIP&44.3&76.1&42.0&83.6&41.1&78.6&34.8&83.0&39.6&76.7&38.8&75.9&40.1&78.9\\
OrdinalCLIP
&45.7&77.6&43.3&85.1&37.9&79.3&36.2&80.4&41.8&77.7&39.5&76.6&40.7&79.4\\
CLIP-DR &\textbf{46.3}&\textbf{83.3}&\textbf{45.8}&\textbf{89.9}&\textbf{48.0}&\textbf{80.7}&\textbf{41.9}&\textbf{84.5}&\textbf{47.3}&\textbf{79.0}&\textbf{41.0}&\textbf{78.9}&\textbf{45.5}&\textbf{82.7}\\
\hline
\end{tabular}%
}
\label{ab_data_pre}
\end{table*}
To assess the effectiveness of CLIP-DR, we conducted a thorough ablation study within the DG test setting and illustrated the AUC scores attained by various models in Table \ref{ab_data_pre}. We evaluated the performance of the model by removing the $\mathcal{L}_{\text{main}}$, the $\mathcal{L}_{\text{rank}}$, and the SMS module, individually, and remaining the rest of the model structure the same. Notably, removing the $\mathcal{L}_\text{{rank}}$ has the most significant impact on AUC (5.5 \%, from an average of 82.7 to 78.4). $\mathcal{L}_{\text{main}}$, $\mathcal{L}_{\text{rank}}$, and SMS module all contribute to the improvement of the DR grading effect, such as 1.3 \%, 5.5\% and 2 \% of AUC score. This experiment demonstrates the importance of natural order information and the effectiveness of our proposed CLIP-DR.

\section{Conclusion}
%In this paper, we further improve the performance of the CLIP model in DR grading by introducing the natural order information of DR images. 
%We propose a novel framework, CLIP-DR, which consists of CLIP baseline, rank loss, and FDS module. while rank loss helps the model to understand the natural order of image-text pairs, the FDS module can minimize the effect of data imbalance. 
%CLIP-DR achieves excellent performance in the publicly available DG test of GDRBench, proving its effectiveness in solving DR grading. The generality of our work enables the core ideas of our work to be applied to many grading tasks with natural order in the medical field.
We propose a novel framework for DR grading with colour fundus images.
It harnesses a ranking-aware prompting strategy to boost the vision-language model's performance by exploiting the natural ordinal information of DR image-text pairs.
Extensive experiments have demonstrated that our CLIP-DR achieves state-of-the-art DR grading performance on an average of six large-scale datasets of the generalizable diabetic retinopathy grading benchmark.

\begin{credits}

\subsubsection{\discintname}
The authors have no competing interests to declare that
are relevant to the content of this article.
\end{credits}
\bibliographystyle{unsrt}
\bibliography{Paper-1493}
\newpage
% This is samplepaper.tex, a sample chapter demonstrating the
% LLNCS macro package for Springer Computer Science proceedings;
% Version 2.20 of 2017/10/04
%
% \documentclass[runningheads]{llncs}
% %
% \usepackage{graphicx}
% \usepackage{comment}
% \usepackage{amsmath,amssymb} % define this before the line numbering.
% \usepackage{color}
% \usepackage{paralist}
% \usepackage{subfigure}
% %\usepackage{subcaption}
% \usepackage{stackengine}
% \usepackage{adjustbox}
% \usepackage{caption} 
% %\captionsetup[table]{skip=5pt}
% %\usepackage{ruler}
% \usepackage{multirow}
% \usepackage{diagbox}
% \usepackage{bbding}
% %\setlength{\textfloatsep}{3pt}
% \usepackage[colorlinks,linkcolor=blue]{hyperref}
% %\usepackage{tikz}
% %\renewcommand\UrlFont{\color{blue}\rmfamily}

% \begin{document}
%
\title{Appendix for ``CLIP-DR: Textual Knowledge-Guided Diabetic Retinopathy Grading with Ranking-aware Prompting"}
\titlerunning{CLIP-DR for Diabetic Retinopathy Grading}
\author{Qinkai Yu \inst{1} \and
Jianyang Xie\inst{2} \and
Anh Nguyen\inst{3} \and
He Zhao\inst{2} \and
Jiong Zhang\inst{4} \and
Huazhu Fu\inst{5} \and
Yitian Zhao\inst{4} \and
Yalin Zheng\inst{2} \and
Yanda Meng \inst{1 (}\Envelope\inst{)}}
\authorrunning{Q.Yu et al.}
\institute{\textsuperscript{$1$} Computer Science Department, University of Exeter, Exeter, UK\\
\textsuperscript{$2$} Eye and Vision Sciences Department, University of Liverpool, Liverpool, UK\\
\textsuperscript{$3$} Computer Science Department, University of Liverpool, Liverpool, UK\\
\textsuperscript{$4$} Ningbo Institute of Materials Technology and Engineering, Chinese Academy of Sciences, Ningbo, China \\
\textsuperscript{$5$} Institute of High Performance Computing, Agency for Science, Technology and Research (A*STAR), Singapore. \\
\email{Y.M.Meng@exeter.ac.uk} \\
}
%index{Qinkai Yu}
%index{Jianyang Xie}
%index{Anh Nguyen}
%index{Jiong Zhang}
%index{Huazhu Fu}
%index{Yitian Zhao}
%index{Yalin Zheng}
%index{Yanda Meng}

\maketitle
\begin{table*}[!h]
\caption{Detailed Result of DG test. Quantitative results on AUC performance of Five Class in DG test setting. Our model performs better than other models and has statistical significance, where ours achieved the best AUC.}
\centering
\resizebox{0.85\textwidth}{!}{%
\begin{tabular}{l|c|c|c|c|c|c|c}
\hline
\multicolumn{8}{c}{Class Normal’s AUC performance}\\
\hline
\multirow{1}{*}{Source} & \multicolumn{1}{|c|}{APTOS} & \multicolumn{1}{|c|}{DeepDR} & \multicolumn{1}{|c|}{FGADR} &\multicolumn{1}{|c|}{IDRID}&\multicolumn{1}{|c|}{Messidor}&\multicolumn{1}{|c|}{RLDR}&\multicolumn{1}{|c}{Average}\\
\hline
GDRNet& 94.3&\underline{88.7}&\underline{87.6}&\underline{85.7}&\textbf{73.6}&71.4&\underline{83.55}\\
\hline
OrdinalCLIP&96.5&86.65&85.93&82.9&71.59&\textbf{77.5}&83.51\\
CLIP-DR &\textbf{97.51}&\textbf{94.25}&\textbf{89.89}&\underline{90.07}&\underline{73.16}&\underline{75.8}&\textbf{86.78}\\
\hline
\multicolumn{8}{c}{Class Mild’s AUC performance}\\
\hline
\multirow{1}{*}{Source} & \multicolumn{1}{|c|}{APTOS} & \multicolumn{1}{|c|}{DeepDR} & \multicolumn{1}{|c|}{FGADR} &\multicolumn{1}{|c|}{IDRID}&\multicolumn{1}{|c|}{Messidor}&\multicolumn{1}{|c|}{RLDR}&\multicolumn{1}{|c}{Average}\\
%\cmidrule(lr){2-6}
%\cmidrule(lr){7-7}
%\cmidrule(lr){8-9}
\hline
GDRNet& \textbf{52.5}&\underline{69.3}&\underline{81.15}&\textbf{82.2}&\underline{50.5}&52.3&\underline{64.6}\\
\hline
OrdinalCLIP&37.79&67.76&79.15&67.5&49.8&\underline{57.6}&59.93\\
CLIP-DR &\underline{51.92}&\textbf{76.48}&\textbf{84.73}&\underline{74.69}&\textbf{52.66}&\textbf{72.79}&\textbf{68.87}\\
\hline
\multicolumn{8}{c}{Class Moderate’s AUC performance}\\
\hline
\multirow{1}{*}{Source} & \multicolumn{1}{|c|}{APTOS} & \multicolumn{1}{|c|}{DeepDR} & \multicolumn{1}{c}{FGADR} &\multicolumn{1}{|c|}{IDRID}&\multicolumn{1}{|c|}{Messidor}&\multicolumn{1}{|c|}{RLDR}&\multicolumn{1}{|c}{Average}\\
%\cmidrule(lr){2-6}
%\cmidrule(lr){7-7}
%\cmidrule(lr){8-9}
\hline
GDRNet&67.2&\textbf{90.8}&\textbf{67.7}&\textbf{76.7}&\textbf{83.3}&\underline{73.1}&\textbf{76.46}\\
\hline
OrdinalCLIP&\underline{74.53}&85.52&48.68&66.28&75.14&\textbf{77.0}&71.19\\
CLIP-DR &\textbf{84.37}&\underline{89.85}&\underline{56.5}&\underline{71.56}&\underline{77.81}&67.6&\underline{73.615}\\
\hline
\multicolumn{8}{c}{Class Severe’s AUC performance}\\
\hline
\multirow{1}{*}{Source} & \multicolumn{1}{|c|}{APTOS} & \multicolumn{1}{|c|}{DeepDR} & \multicolumn{1}{|c|}{FGADR} &\multicolumn{1}{|c|}{IDRID}&\multicolumn{1}{|c|}{Messidor}&\multicolumn{1}{|c|}{RLDR}&\multicolumn{1}{|c}{Average}\\
%\cmidrule(lr){2-6}
%\cmidrule(lr){7-7}
%\cmidrule(lr){8-9}
\hline
GDRNet&83.3&81.9&71.5&\textbf{83.3}&\textbf{95.4}&79.4&82.4\\
\hline
OrdinalCLIP&\underline{85.83}&\underline{90.65}&\textbf{76.17}&\underline{82.7}&89.6&\textbf{84.9}&\textbf{84.975}\\
CLIP-DR &\textbf{89.04}&\textbf{91.26}&\underline{69.2}&80.88&\underline{92.6}&\underline{83.80}&\underline{84.46}\\
\hline
\multicolumn{8}{c}{Class Proliferative’s AUC performance}\\
\hline
\multirow{1}{*}{Source} & \multicolumn{1}{|c|}{APTOS} & \multicolumn{1}{|c|}{DeepDR} & \multicolumn{1}{|c|}{FGADR} &\multicolumn{1}{|c|}{IDRID}&\multicolumn{1}{|c|}{Messidor}&\multicolumn{1}{|c|}{RLDR}&\multicolumn{1}{|c}{Average}\\
%\cmidrule(lr){2-6}
%\cmidrule(lr){7-7}
%\cmidrule(lr){8-9}
\hline
GDRNet&84.7&85.8&\textbf{76.0}&\textbf{92.8}&95.4&\textbf{89.0}&\underline{87.2}\\
\hline
OrdinalCLIP&\underline{84.77}&\textbf{91.51}&69.44&86.6&99.2&84.2&85.95\\
CLIP-DR &\textbf{91.94}&\underline{90.15}&\underline{70.79}&\underline{90.68}&\textbf{99.3}&\underline{85.69}&\textbf{88.1}\\
\hline
\end{tabular}
}
\label{Class1}
\end{table*}
\begin{figure*}[tb]
\centering
{\includegraphics[width=0.85\linewidth]{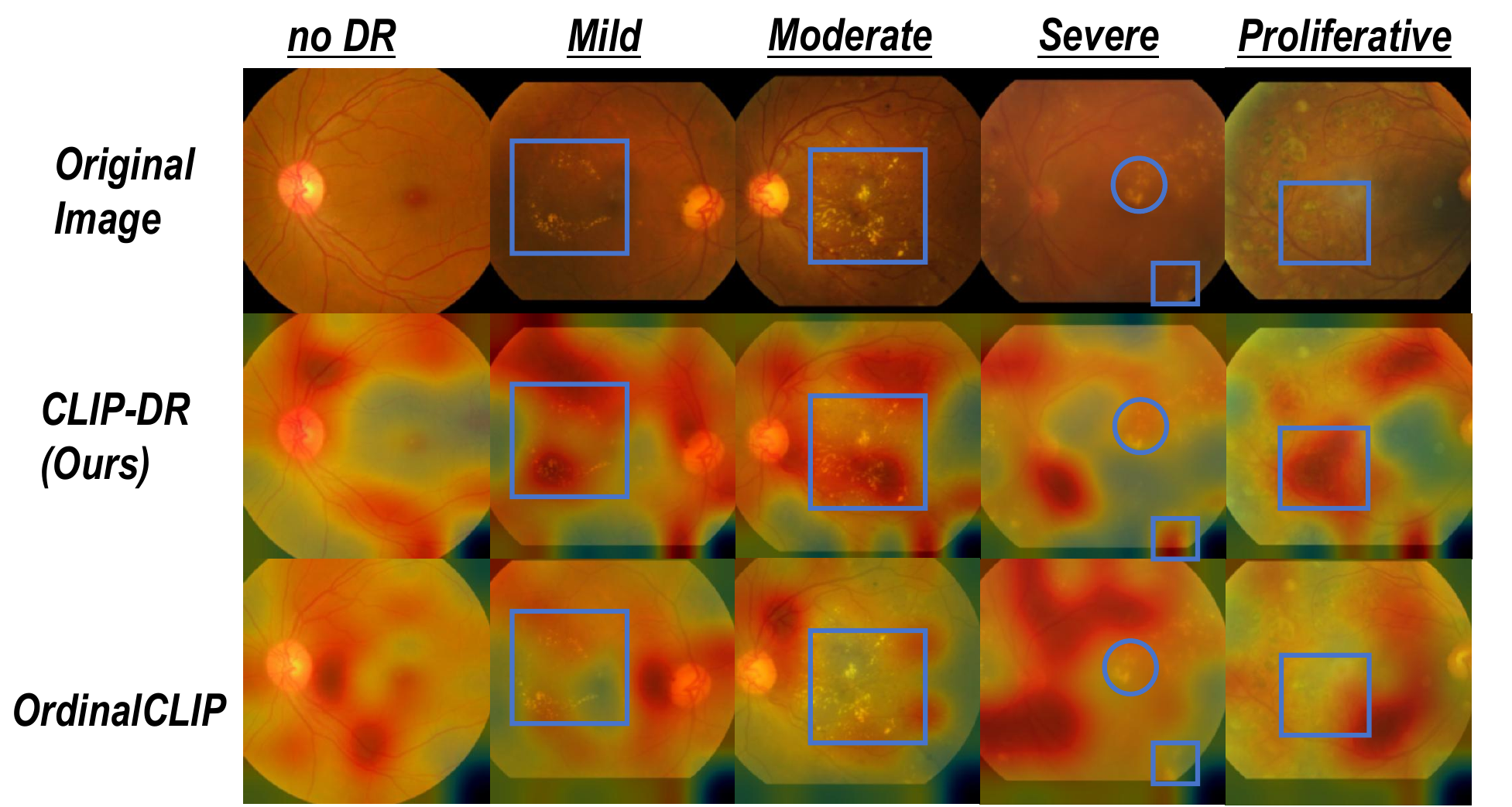}}
\caption{Class Activation Map for CLIP-DR and OrdinalCLIP. The top line is the original image, the second line is CLIP-DR (Ours), and the last line is OrdinalCLIP. CLIP-DR and OrdinalCLIP use the same training data (DG test, `APTOS' as target)and train 100 epochs. We highlighted the differences in class activation diagrams with blue boxes.}
\label{cam}
\end{figure*}
\begin{table*}[!h]
\caption{Comparison with state-of-the-art approaches under the ESDG Test setting. The method, such as GDRNet, mainly focuses on the finite-domain transfer problem; thus, it fits the ESDG testing set; however, our proposed CLIP-DR does not apply additional special tricks to boost the performance. For the ESDG test setting, CLIP, OrdinalCLIP, and our CLIP-DR are all fine-tuned on a very small dataset (the ratio of the training set to the test set is about 1:50) and thus do not achieve the performance comparable to the state-of-the-art method GDRNet. Our CLIP-DR achieves the best average AUC and F1, compared with CLIP and OrdinalCLIP.
% Comparison with state-of-the-art approaches under the ESDG Test setting. The method, such as GDRNet, mainly focuses on the finite-domain transfer problem; thus, it fits the ESDG testing set; however, our proposed CLIP-DR does not apply additional special tricks to boost the performance. For the ESDG test setting, CLIP, OrdinalCLIP, and our CLIP-DR are all fine-tuned on a very small dataset (the ratio of the training set to the test set is about 1:50) and thus do not achieve the performance comparable to the state-of-the-art method GDRNet. However, a relatively lower performance for all the compared methods on average, such as 20-30\% F1 score, can demonstrate the unsuitability of such an experimental setting. Our CLIP-DR achieves the best average AUC and F1, compared with CLIP and OrdinalCLIP.
}
\renewcommand\arraystretch{1.5}
\centering
\resizebox{0.85\textwidth}{!}{%
\begin{tabular}{l|cc|cc|cc|cc|cc|cc|cc}
\hline
\multirow{1}{*}{Source} & \multicolumn{2}{|c|}{APTOS} & \multicolumn{2}{|c|}{DeepDR} & \multicolumn{2}{|c|}{FGADR} &\multicolumn{2}{|c|}{IDRID}&\multicolumn{2}{|c|}{Messidor}&\multicolumn{2}{|c|}{RLDR}&\multicolumn{2}{|c}{Average}\\
%\cmidrule(lr){2-6}
%\cmidrule(lr){7-7}
%\cmidrule(lr){8-9}
\hline

Metrics& AUC  & F1  & AUC & F1 & AUC &F1&AUC&F1&AUC&F1&AUC&F1&AUC&F1 \\
\hline
\hline

DRGen& 69.4  & \textbf{35.7} & \textbf{78.5} & 31.6 & 59.8 &8.4&70.8&30.6&\underline{77.0}&\underline{37.4}&78.9&21.2&\underline{72.4}&27.5  \\
Mixup& 65.5 &30.2 &70.7 &33.3 &58.8 &7.4&70.2&32.6&71.5&32.6&72.9&27.0&68.3&27.2 \\
MixStyle& 62.0  & 25.0 & 53.3  & 14.6 & 51.0 &7.9&53.0&19.3&51.4&16.8&53.5&6.4&54.0&15.0 \\
GREEN& 67.5  & 33.3 & 71.2 & 31.1 &58.1 &6.9&68.5&33.0&71.3&33.1&71.0&27.8&67.9&27.5 \\
CABNet& 67.3  & 30.8 & 70.0 & 32.0 & 57.1 &7.5&67.4&31.7&72.3&35.3&75.2&25.4&68.2&27.2 \\
DDAIG&67.4 &31.6 & 73.2  & 29.7 & 59.9 &5.5&70.2&33.4&73.5&35.6&74.4&23.5&69.8&26.7 \\
ATS &68.8&32.4&72.7&33.5&60.3&5.7&69.1&30.6& 73.4&32.4&75.0&23.9&69.9&26.4\\
Fishr&64.5&31.0&72.1&30.1&56.3&7.2&71.8&30.6&74.3&33.8&\underline{78.6}&21.3&69.6&25.7\\
MDLT& 67.6&32.4&73.1&33.7&57.1&7.8&\underline{71.9}&32.4&73.4&34.1&76.6&\underline{30.0}&70.0&\underline{28.4}\\

GDRNet& \underline{69.8}&\underline{35.2}&\underline{76.1}&\textbf{35.0}&\textbf{63.7}&\underline{9.2}&\textbf{72.9}&\textbf{35.1}&\textbf{78.1}&\textbf{40.5}&\textbf{79.7}&\textbf{37.9}&\textbf{73.4}&\textbf{32.2}\\
\hline
CLIP&60.3 &28.0 &64.7 &27.1 &59.0 &7.7 &59.8 &21.8 &65.4 &27.7 &69.3 &25.6 &63.1 &23.0 \\
OrdinalCLIP&61.6&27.4&66.9&28.4&60.4&8.1&61.6&23.8&67.6&29.5&70.1&26.3&64.7&23.9\\
CLIP-DR(Ours)&\textbf{69.8}&30.5&72.8&31.0&\underline{60.5}&\textbf{9.7}&66.6&27.9&68.9&31.2&72.3&27.7&68.5&26.3\\
\hline
\end{tabular}%
}
\label{ESDG}
\end{table*}

\end{document}